\documentclass[a4paper,11pt]{article}
\usepackage{amssymb,amsmath,amsthm,amsfonts}
\usepackage{bookmark}
\usepackage{mathrsfs}
\usepackage{multirow}
\usepackage{graphicx}
\usepackage{authblk}
\usepackage{indentfirst}
\usepackage{multicol}
\usepackage{tabu}
\usepackage{tabularx}
\usepackage{url}
\usepackage{fancyhdr}
\usepackage[numbers]{natbib}
\usepackage[all]{xy}
\usepackage{mathtools}
\usepackage[english]{babel}
\usepackage{colortbl}
\usepackage{caption}
\usepackage{hyperref}
\usepackage{xcolor}
\usepackage[mathlines]{lineno}

\usepackage{xr}
\makeatletter
    \newcommand*{\addFileDependency}[1]{
    \typeout{(#1)}
    \@addtofilelist{#1}
    \IfFileExists{#1}{}{\typeout{No file #1.}}
    }
\makeatother

\title{Interaction Topological Transformer for Multiscale Learning in Porous Materials}

\author[1]{Dong Chen} 
\author[1]{Jian Liu} 
\author[2]{Chun-Long Chen \thanks{Corresponding author: chunlong.chen@pnnl.gov}}
\author[1,3,4]{Guo-Wei Wei \thanks{Corresponding author: weig@msu.edu}}
\affil[1]{Department of Mathematics, Michigan State University, MI, 48824, USA}
\affil[2]{Physical Sciences Division, Pacific Northwest National Laboratory, Richland, Washington 99354, United States}
\affil[3]{Department of Electrical and Computer Engineering, Michigan State University, MI 48824, USA}
\affil[4]{Department of Biochemistry and Molecular Biology, Michigan State University, MI 48824, USA}

\makeatletter
    \renewcommand*{\@fnsymbol}[1]{\ensuremath{\ifcase#1\or \dagger\or *\or *\or
   \mathsection\or \else\@ctrerr\fi}}
\makeatother
\date{}

\begin{document}
    \maketitle

    \paragraph{Abstract}
Porous materials exhibit vast structural diversity and support critical applications in gas storage, separations, and catalysis. However, predictive modeling remains challenging due to the multiscale nature of structure–property relationships, where performance is governed by both local chemical environments and global pore-network topology. These complexities, combined with sparse and unevenly distributed labeled data, hinder generalization across material families. We propose the Interaction Topological Transformer (ITT), a unified  data-efficient framework that leverages novel interaction topology to capture materials information across multiple scales and multiple levels, including structural, elemental, atomic, and pairwise-elemental organization. ITT extracts scale-aware features that reflect both compositional and relational structure within complex porous frameworks, and integrates them through a built-in Transformer architecture that supports joint reasoning across scales. Trained using a two-stage strategy, i.e., self-supervised pretraining on 0.6 million unlabeled structures followed by supervised fine-tuning, ITT achieves state-of-the-art, accurate, and transferable predictions for adsorption, transport, and stability properties. This framework provides a principled and scalable path for learning-guided discovery in structurally and chemically diverse porous materials.


    \newpage
    \tableofcontents
    \newpage

\section{Introduction}\label{section:introduction}

Porous materials such as metal–organic frameworks (MOFs), covalent organic frameworks (COFs), zeolites, and porous polymers underpin applications in gas storage and separations \cite{chen2022porous}, carbon capture \cite{siegelman2021porous}, catalysis \cite{li2024recent}, and sensing \cite{sharma2024porous,tao2024resistive}. Their technological promise arises from modular building blocks, tunable pore architectures, and rich surface chemistries, which together create an effectively unbounded design space with many structures already synthesized and many more accessible in silico \cite{colon2014high,lee2021computational}. For example, polyoxometalate-based MOFs (POM-MOFs) represent a particularly versatile class, combining the tunable electronic and redox properties of POMs \cite{yang2023old} with the open, modular networks of MOFs. These hybrid materials have shown great potential in heterogeneous catalysis, where the POM units provide functional activity while the MOF framework ensures accessibility and structural tunability \cite{ebrahimi2024polyoxometalate}, exemplifying how functional building blocks and networked architectures can be integrated to advance porous-material applications \cite{du2014recent}.

However, accurate prediction remains difficult because performance depends not only on local surface chemistry but also on the long-range topology and connectivity of the pore network, which complicates faithful and efficient simulation across conditions and tasks; conducting independent virtual screenings for each application is impractical at scale \cite{daglar2020recent,yadav2024influence}.
More broadly, property prediction in porous materials remains challenging because structure–property relationships are inherently multiscale: behavior emerges from the interplay between local chemical environments and the global geometry and connectivity of pore networks. Performance is also highly sensitive to operating conditions, such as temperature, pressure, and adsorbate identity, with long-range, network-mediated effects often outweighing purely local descriptors \cite{morris2017coordination,bennett2021changing}. Compounding these issues, labeled data are limited and uneven across material families and target properties, and are generated under diverse experimental and computational protocols, which complicates supervised learning and hinders generalization \cite{jablonka2020big}.
These realities motivate learning approaches that integrate local chemistry with pore-network topology while remaining data-efficient and transferable across porous-material families and operating conditions.

Prior work spans physics-informed descriptors \cite{rosen2021machine,shi2023two}, such as surface area, pore-size distributions, and chemical fingerprints, as well as learned representations based on graph and geometric neural networks (GNNs) originally developed for molecules and dense crystals \cite{xie2018crystal,chen2019graph}. In parallel, math-based approaches \cite{krishnapriyan2021machine,chen2025category,chen2025enhancing} have been explored to incorporate mathematical principles into porous materials modeling. While these methods face limitations in porous-materials settings: hand-crafted descriptors only partially encode void topology and channel connectivity; local message passing under-represents higher-order (multi-body) interactions \cite{chen2025enhancing} and long-range effects transmitted through connected pores; and family-specific models transfer poorly across MOFs, COFs, zeolites, and porous polymers \cite{park2023enhancing}. In practice, realistic frameworks often require multiple periodic images or large supercells—frequently exceeding $10^4$ atoms—to capture pore networks and guest–framework interactions, rendering conventional GNNs computationally prohibitive and memory-bound. Voxelization and other 3D-grid approaches convert structures into images to leverage computer-vision architectures \cite{cho2021nanoporous}, but they sacrifice key invariances (rotation, translation, permutation), suffer from extreme sparsity and dimensionality, and complicate faithful treatment of periodicity. Moreover, existing pretraining strategies commonly rely on engineered priors \cite{park2023enhancing,kang2023multi}, such as connectivity types or hand-computed void metrics, as supervision, which limits scalability and can bias the learned representation rather than allowing the model to discover transferable topological structure directly from raw data.

Here, we introduce interaction topology \cite{liu2024persistent,liu2023interaction} as a novel multilevel interactive representation that integrates structural, elemental, and pairwise elemental information within a single formalism. Unlike classical persistent homology (PH) \cite{zomorodian2004computing}, which analyzes a system as a whole, interaction topology examines how element-defined subsets contribute both individually and jointly. We instantiate this via an interaction complex (IC), a generalization of the simplicial complex that encodes higher-order and element-specific information. Applying persistent interaction homology (PIH) to this complex yields scale-aware summaries at three complementary levels: form individual clusters, cluster interactions, to global pore architecture.   
In parallel, we preserve an atomic view: a Crystal Graph Convolutional Neural Network (CGCNN) \cite{xie2018crystal} module (without global pooling) produces locality-preserving atomic embeddings, ensuring that fine-scale chemical and geometric signals are retained rather than collapsed. Building on these ingredients, the Interaction Topological Transformer (ITT) fuses four embedding streams, including structural, elemental, atomic, and pairwise interaction, into a unified model. A transformer attends across all tokens so that topology-derived patterns can reinforce or counterbalance localized chemical environments. Cross-attention layers inject interaction-topology signals into the multilevel pathway, encouraging the network to integrate long-range pore-network dependencies with local information. The CGCNN branch refines atomic tokens before fusion, while the topology channels contribute persistence-based descriptors of connected components, tunnels, and cavities across scales. After multiple ITT layers, property-specific output heads support adsorption, transport, and stability queries within the ITT architecture.

Furthermore, we adopt a two-stage training strategy. During pretraining, ITT is exposed to a large unlabeled corpus (over 0.6 million structures) spanning MOFs, COFs, zeolites, porous polymers, and other pore-network materials—without external priors, auxiliary data, or labels. Self-supervised learning recovers masked structural-level embeddings from the remaining multilevel context (atomic, elemental, and interaction-topology signals), encouraging alignment between geometric, topological, and chemical views. In fine-tuning, the pretrained backbone is adapted to labeled benchmarks covering a variety of properties—including adsorption (Henry coefficients, uptakes, selectivities)\cite{banerjee2016metal,orhan2021prediction}, transport (diffusivities)\cite{daglar2022combining}, and stability/mechanical responses \cite{nandy2021using}. Across these benchmarks, ITT delivers accurate, transferable, and family-agnostic predictions, outperforming previous baselines and improving cross-family generalization. The multilevel design also provides interpretability: attention concentrated on high-persistence structural and interaction features aligns with known adsorption sites and rate-limiting apertures, while atom-level signals remain traceable through the CGCNN branch. Taken together, the proposed interaction topological Transformer forms a coherent framework that reconciles local chemistry with pore-network topology for data-efficient prediction across diverse porous materials.

\section{Results}

\subsection{Interaction Topological Transformer}

  \begin{figure}[!ht]
    \centering
    \includegraphics[width=16cm]{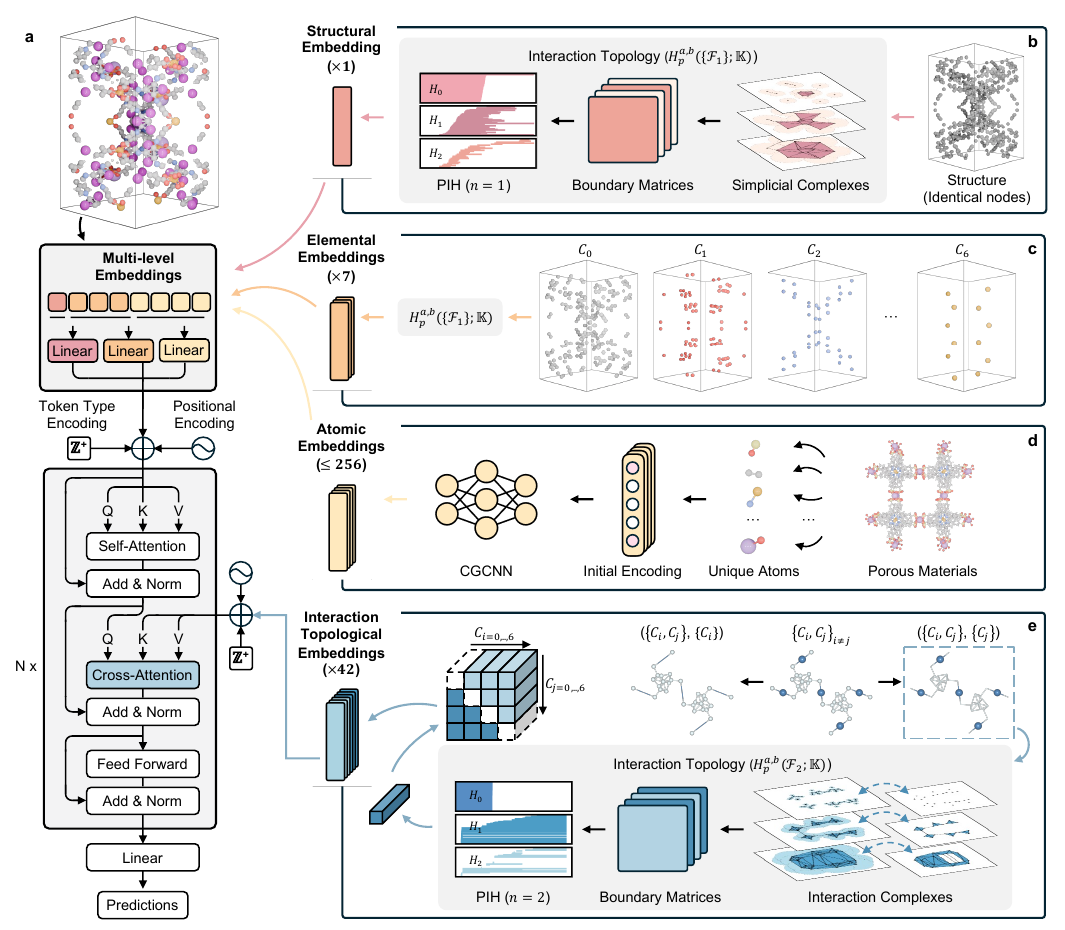}
    \caption{Overall framework of the Interaction Topological Transformer (ITT).
{\bf a} Given a porous material structure, ITT constructs multi-level embeddings consisting of structural-, elemental-, atomic-, and interaction-topological–level representations. 
{\bf b} Structural embedding: the entire structure is modeled as simplicial complexes where all atoms are treated as identical nodes. Persistent interaction homology (PIH, $n = 1$) is applied to extract topological features, yielding a single structural token. 
{\bf c} Elemental embeddings: atoms are grouped into seven element clusters ($C_0$–$C_6$), and PIH ($n = 1$) is computed for each cluster, producing seven tokens that represent cluster-specific topological contributions.
{\bf d} Atomic embeddings: a CGCNN without pooling encodes unique atoms, generating locality-preserving atomic representations with up to 256 atomic tokens. 
{\bf e} Interaction topological embeddings: element-cluster interactions are represented by interaction simplicial complexes, where PIH ($n = 2$) captures higher-order relations across clusters. Considering 7 element clusters, 42 pairwise interaction tokens are derived. Together, these multi-level embeddings are fused through self-attention and cross-attention layers to form the final predictive representation of porous material properties.
    }
    \label{fig:framework}
  \end{figure}

A central contribution of the proposed ITT framework is its integration of persistent interaction homology as a topological foundation for multilevel material representation. While traditional graph-based models are limited to pairwise atomic interactions, and classical PH focuses on global topology without differentiating atom types, PIH advances this landscape by introducing interaction topology-a formalism designed to capture structural, elemental, and cross-elemental organization within a unified framework \cite{liu2024persistent,liu2023interaction}. At its core, PIH operates on interaction complexes (ICs), which are a generalizations of simplicial complexes that encode both intra- and inter-cluster geometric relationships. By applying PIH to these complexes, the ITT model generates scale-aware descriptors at three complementary levels: (i) a structural-level embedding summarizing the global pore geometry; (ii) elemental-level embeddings based on clusters derived from co-occurrence and chemical similarity; and (iii) pairwise interaction-level embeddings that characterize the topological relationships between elemental clusters (e.g., S–Tb interactions). Figure~\ref{fig:discussion_analysis}{\bf c–f} illustrate how PIH captures both conventional features (e.g., connected components, tunnels, voids) and emergent, chemically informed interaction motifs that classical PH would overlook.

Importantly, PIH enables the model to retain local chemical detail while also learning from inter-element interactions, a capability demonstrated in the comparison of different interaction centers in Figure~\ref{fig:discussion_analysis}{\bf e–f}. For example, choosing S versus Tb as the interaction center yields qualitatively different $H_0$ and $H_1$ topological patterns—capturing, respectively, the number of centers and the multi-scale formation of triangle-like structures. Through multiscale filtrations (e.g., $\alpha$ = 1, 2, 2.5 \AA), PIH systematically quantifies the evolution of interaction geometry, preserving both fine-grained local features and global connectivity. This interaction-centered perspective is critical for modeling porous materials, where function often emerges from the joint behavior of chemically distinct subsystems.

Building upon this topological foundation, ITT fuses four embedding streams—structural, elemental, atomic, and pairwise interaction—into a unified Transformer architecture. The topology-derived embeddings are projected and fed as discrete tokens into the Transformer, where cross-attention layers inject interaction-topology signals into the attention mechanism, encouraging alignment between pore-level organization and atomic-scale environments. The transformer layers allow these representations to inform one another: high-persistence topological features can reinforce or modulate local cues, and vice versa. This design enables ITT to model both local chemistry and long-range pore-network topology, making it highly suitable for data-efficient, multi-property prediction.

Furthermore, ITT adopts a two-stage training strategy. During self-supervised pretraining, the model is exposed to over 0.6 million unlabeled porous structures spanning MOFs, COFs, ZEOs, PPNs, and others. The model learns to recover masked structural-level embeddings from the context of elemental, atomic, and interaction-topological embeddings—thus enforcing consistency between topological and chemical views without requiring external labels or handcrafted features. In the fine-tuning stage, ITT is adapted to supervised benchmarks across a wide range of properties, including adsorption (e.g., Henry coefficients), transport (e.g., diffusivity), and stability (e.g., mechanical resilience). Across these tasks, ITT achieves strong predictive performance and cross-family generalization, outperforming conventional baselines.

Finally, the multilevel design of ITT supports interpretability. Attention maps reveal that high-persistence structural and interaction tokens often align with physically meaningful regions—such as known adsorption sites or bottleneck apertures—while atomic-level contributions remain traceable through the CGCNN branch. Taken together, PIH plays a central role in enabling ITT to reconcile local chemical signals with global and relational topological features, forming a coherent and versatile architecture for learning from complex porous materials. Looking forward, the integration of interpretable topological and chemical reasoning in ITT holds strong potential for guiding the design of functional porous materials, accelerating materials discovery, and enabling robust, scalable modeling of large and structurally diverse frameworks encountered in real-world applications.

\subsection{Schematic Workflow}

  \begin{figure}[!ht]
    \centering
    \includegraphics[width=16cm]{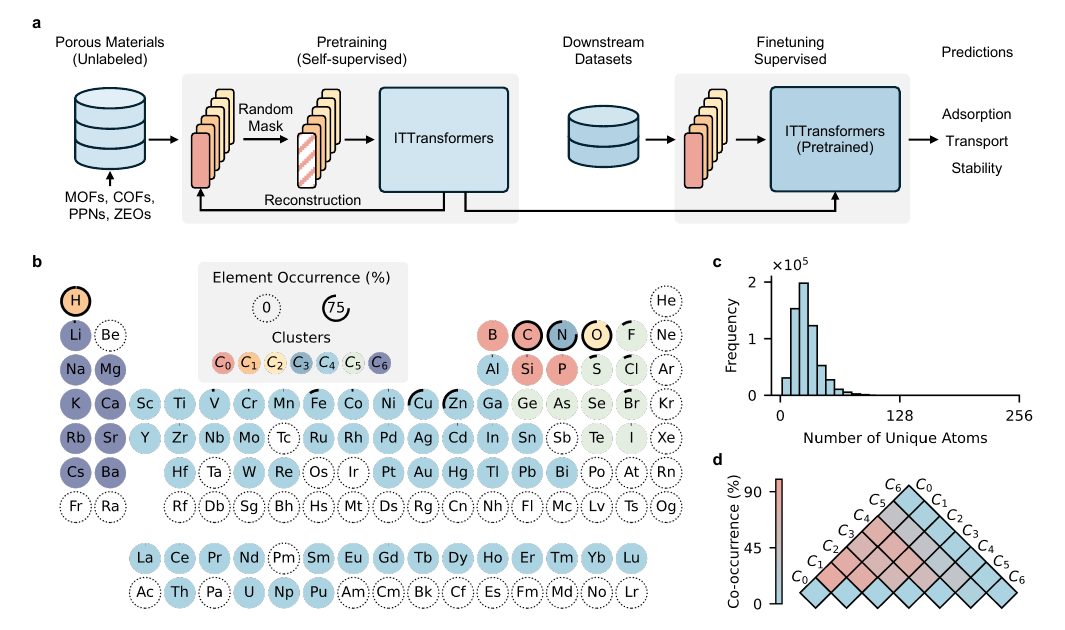}
    \caption{Workflow and pre-analysis of the pretraining dataset.
{\bf a} Overall workflow: Unlabeled porous materials, including MOFs, COFs, PPNs, IZAs, are used for self-supervised pretraining, where random masking for structural embedding and reconstruction tasks are performed with ITT. The pretrained model is then fine-tuned on downstream datasets to predict adsorption, transport, and stability properties.  
{\bf b} Elements in porous materials are grouped into seven clusters ($C_0$–$C_6$) by considering both co-occurrence statistics and chemical similarity. The size of each circle reflects element occurrence frequency, with H, C, N, and O dominating. 
{\bf c} Unique atom distribution: Histogram of the number of unique atoms per structure across the pretraining dataset.
{\bf d} Pairwise co-occurrence percentages among the seven element clusters show strong correlations between different clusters but weak self-co-occurrence within the same cluster.
    }
    \label{fig:workflow}
  \end{figure}

The schematic workflow of the proposed approach is illustrated in Figure~\ref{fig:workflow}{\bf a}. To leverage the abundance of available structural data, we first collected a large and diverse data of porous material structures, without requiring any labeled property information, including MOFs, COFs, PPNs, and IZAs. In total, approximately 0.6 million structures were assembled for pretraining. Detailed sources and statistics for the collected data are provided in Table~\ref{tab:datasets} and discussed in the Method section. 
Following data collection, we introduced the ITT model (Figure~\ref{fig:framework}) to perform self-supervised pretraining. Each structure is first processed to generate multi-level embeddings. During pretraining, a subset of the structure-level embeddings is randomly masked, and the model is trained to reconstruct the masked portion based on the remaining multi-level context. This self-supervised objective requires no external labels, in contrast to previous supervised methods, and motivates the model to learn the underlying relationships of multi-level patterns across materials. By learning from intrinsic structural organization alone, the pretrained ITT becomes capable of modeling previously unseen porous materials \cite{kang2023multi}. In the fine-tuning stage, the same embedding procedure is applied to structures in labeled downstream datasets, and the pretrained ITT is adapted via supervised learning to predict specific properties. These downstream tasks span adsorption, transport, and stability domains. The two-stage learning paradigm, which is self-supervised pretraining followed by task-specific fine-tuning, allows ITT to generalize effectively even in low-data regimes, while capturing both local and global structural signals critical to property prediction.

\subsection{Data Preprocessing Rationale}\label{sec:preanalysis}
Porous materials, especially MOFs, exhibit high elemental diversity, spanning over 70 elements in the periodic table. While the global dataset includes a vast variety of elements, each individual structure typically contains only a limited subset, which often just a few unique atomic environments. This imbalance poses a significant challenge for machine learning models: representing each element independently leads to sparse, high-dimensional inputs that are both inefficient and difficult to generalize from. Therefore, it is essential to develop a compact and chemically meaningful element representation that captures relevant distinctions without unnecessary complexity.

To address this, we performed a pre-analysis of element usage across approximately 0.6 million porous structures, with the goal of clustering elements into a small number of groups that preserve both chemical relevance and statistical interpretability. This clustering not only reduces input dimensionality but also enables robust modeling of inter-cluster interactions, which are central to our proposed ITT framework. The clustering strategy integrates two principles: (1) Co-occurrence statistics. Elements that frequently appear together are likely functionally distinct and should be separated to preserve interaction information. (2) Chemical similarity. Elements with similar properties are more likely to play analogous roles across materials and can be grouped accordingly. Specifically, to avoid biasing the final clustering results due to the data collection process, the verified and diverse ACR-MOF \cite{burner2023arc} dataset was used to determine the choice of clusters.

The strategy begins by filtering out rarely used or chemically unstable elements (atomic number $>$ 103). A co-occurrence matrix is constructed to capture how often each element pair appears across structures. To avoid clustering highly co-occurring elements (which may be functionally complementary), we invert and normalize this matrix to serve as a dissimilarity signal. In parallel, we extract chemically meaningful features from the periodic table, i.e., atomic number, group, block, electronegativity, valence electrons, and atomic radius, and compute a cosine similarity matrix representing fundamental chemical likenesses. Next, we define anchor elements for initial cluster centers based on frequency and chemical diversity: C, H, O, N, and Zn, which dominate the dataset (top 5 leading elements) and represent core chemical functionalities. The remaining elements are grouped using a hybrid optimization of co-occurrence dissimilarity and chemical similarity, resulting in seven element clusters ( C$_0$– C$_6$). The resulting clustering is visualized in Figure~\ref{fig:workflow}{\bf b}, where black circle length reflects element frequency. As expected, H, C, N, and O dominate across the dataset, motivating their selection as separate clusters. The total count of elements along with their occurrence rates is shown in Supplementary Table S4. Additionally, the distribution of unique atomic environments per structure, shown in Figure~\ref{fig:workflow}{\bf c}, demonstrates that most structures contain far fewer than 256 unique atoms, thereby justifying the choice of 256 as the maximum token number for atomic-level embeddings.

Figure~\ref{fig:workflow}{\bf d} presents the pairwise co-occurrence heatmap between the seven clusters. The high co-occurrence between different clusters and low co-occurrence within the same cluster across all structures in the pretraining dataset supports the validity of the clustering strategy. Moreover, in 93.67\% of the structures, the elements are distinctly assigned to different clusters. This separation ensures that cluster-based abstraction does not obscure important interaction signals and element-level distinctions are preserved at the structural level, enabling accurate interaction modeling in the ITT framework.

\subsection{Results and Analysis of the ITT}

  \begin{figure}[!ht]
    \centering
    \includegraphics[width=16cm]{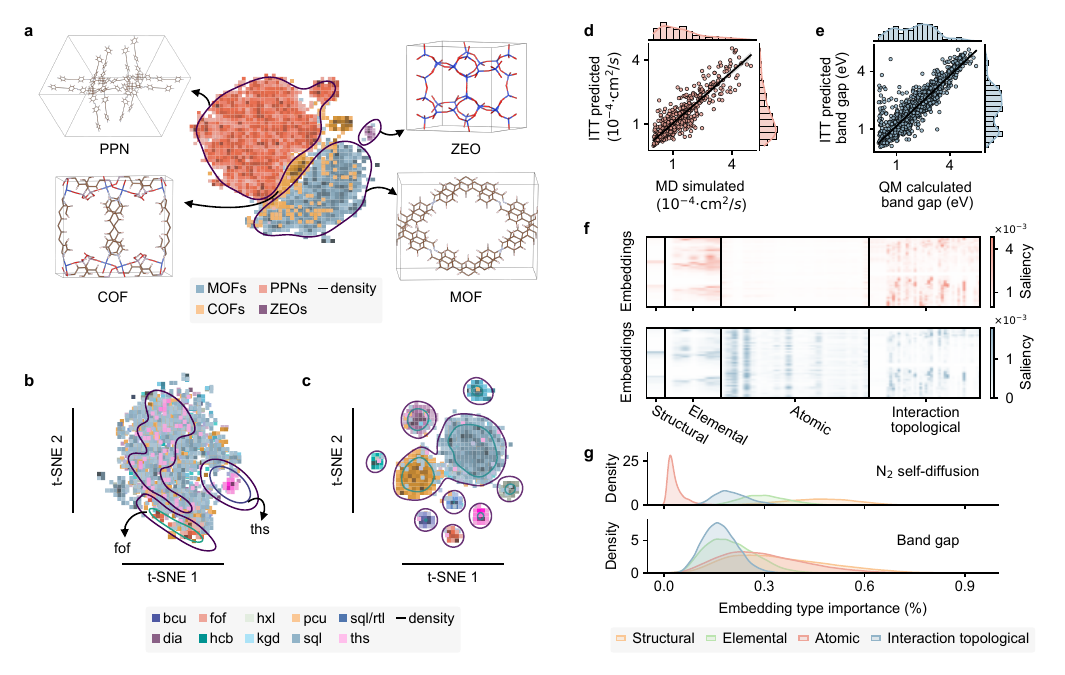}
    \caption{ITT model analysis and visualization of representation learning.
    {\bf a} 2D t-SNE visualization of the latent features (first token embeddings using for prediction) from the pretrained ITT model across four porous material types—MOFs, COFs, PPNs, and ZEOs—colored by class. Density contours (darker regions) indicate higher local concentration of samples, and representative structures are shown for each class.
    {\bf b} t-SNE visualization of pretrained ITT's latent features for the CoRE MOF 2019 dataset, colored by topological class. Highlighted regions show dense clusters for specific topologies, such as ths and fof.
    {\bf c} t-SNE visualization of the finetuned ITT model trained on a multi-class topology classification task using CoRE MOF 2019, showing increased separation and class-specific clustering of topologies.
    {\bf d} Regression comparison between ITT-predicted and MD-simulated self-diffusion coefficients for N$_2$ (MAE$=3.23\times{10}^{-5}$), in units of $10^{-4} \, \text{cm}^2/\text{s}$. Each dot represents one porous structure.
    {\bf e} Comparison between ITT-predicted and QM-calculated band gaps (MAE=0.263 eV), where each dot corresponds to a single material.
    {\bf f} Saliency score heatmaps for ITT models finetuned on N$_2$ self-diffusion (top) and band gap prediction (bottom), showing the contribution of each embedding token to the final prediction. Token groups include structural ($\times 1$), elemental ($\times 7$), atomic ($\leq256$), and interaction topological embeddings ($\times 42$). The MOF (ID: 'ABAXUZ\_FSR') from CoRE MOF 2019 dataset was used to generate the saliency score.
    (g) Distributions of embedding type importance across samples, shown for models finetuned on N$_2$ diffusion (top) and band gap (bottom). The $x$-axis represents the percentage contribution of each embedding type to the prediction, and the y-axis shows the sample density.
    }
    \label{fig:results_analysis}
  \end{figure}

To assess the representational capacity of the pretrained ITT model, we visualize the learned latent features using 2D t-SNE projections. As shown in Figure~\ref{fig:results_analysis}{\bf a}, the embeddings from the first token (prior to the prediction head) exhibit clear clustering by material type, despite being trained without labels. Distinct groupings are observed for well-refined datasets such as MOFs (CoRE MOF 2019 \cite{chung2019advances}), COFs (CoRE-COFs v7 \cite{tong2017exploring}), PPNs \cite{park2024large}, and ZEOs (IZA-SC \cite{baerlocher2007atlas}), suggesting that ITT's self-supervised pretraining is capable of capturing underlying structural and chemical regularities across diverse porous materials. In contrast, when applied to more compositionally and structurally diverse hypothetical datasets, such as hPPNs \cite{martin2014silico} and MC-COFs \cite{mercado2018silico}, the resulting t-SNE embeddings (Supplementary Figure S1) display considerable overlap, indicating that additional supervision may be required to resolve fine-grained distinctions in complex systems. Beyond material-type separation, the pretrained ITT model also encodes topological organization within a single dataset (MOFs). Figure~\ref{fig:results_analysis}{\bf b} shows that structures in the CoRE MOF 2019 dataset cluster according to known topologies (e.g., fof, ths), despite no access to topology labels during training. This demonstrates the model’s ability to learn meaningful topological features directly from structure. Upon finetuning ITT on a supervised classification task involving the 10 most frequent topologies, the latent space becomes more structured and class-discriminative (Figure~\ref{fig:results_analysis}{\bf c}), confirming the model’s adaptability and effectiveness in downstream tasks.

\begin{table}[ht]
\centering
\caption{Performance comparison across downstream porous materials datasets$^a$}
\label{tab:downstream_scratch_vs_itt_mae}
\resizebox{\textwidth}{!}{%
\begin{tabular}{l l l r l l l l}
\hline\hline
\textbf{Data Type} & \textbf{Dataset} & \textbf{Task} & \textbf{Size} & \textbf{Metric} & \textbf{Source} & \textbf{Scratch} & \textbf{ITT} \\
\hline
MOFs & O$_2$/N$_2$-Select & Henry constant (N$_2$)                      & 4744 & R$^2$   & 0.70 \cite{orhan2021prediction} & 0.80             & {\bf 0.82} \\
     &                    & Henry constant (O$_2$)                      & 5036 & R$^2$   & 0.74 \cite{orhan2021prediction} & 0.85             & {\bf 0.86} \\
     &                    & Uptake (N$_2$)                              & 5132 & R$^2$   & 0.71 \cite{orhan2021prediction} & 0.83             & {\bf 0.83} \\
     &                    & Uptake (O$_2$)                              & 5241 & R$^2$   & 0.74 \cite{orhan2021prediction} & 0.85             & {\bf 0.87} \\
     &                    & Self-diffusivity (N$_2$, 1 bar)             & 5056 & R$^2$   & 0.76 \cite{orhan2021prediction} & 0.77             & {\bf 0.81} \\
     &                    & Self-diffusivity (N$_2$, $\infty$ dilution) & 5192 & R$^2$   & 0.76 \cite{orhan2021prediction} & 0.78             & {\bf 0.82} \\
     &                    & Self-diffusivity (O$_2$, 1 bar)             & 5223 & R$^2$   & 0.74 \cite{orhan2021prediction} & 0.71             & {\bf 0.76} \\
     &                    & Self-diffusivity (O$_2$, $\infty$ dilution) & 5097 & R$^2$   & 0.78 \cite{orhan2021prediction} & 0.75             & {\bf 0.79} \\
     & CO$_2$--Henry      & CO$_2$ Henry (log $k_H$)                    & 9525 & MAE     & 4.94e$-$01$^b$                  & 3.38e$-$01       & {\bf 3.20e$-$01} \\
     & QMOF--BD           & Bandgap                                     & 20375& MAE     & 2.69e$-$01$^b$                  & {\bf 2.63e$-$01} & 2.72e$-$01 \\
     & SRS$^c$            & Solvent-removal stability                   & 2179 & Accuracy& {\bf 0.76$^d$} \cite{nandy2021using}      & 0.71$^d$ (0.74)             & 0.70$^d$ (0.77) \\
     & TST$^c$            & Thermal stability                           & 3132 & MAE     & 44\cite{nandy2021using}                    & 46$^d$(50.2)             & {\bf 43}$^d$(48.8) \\
\hline
COFs & MC--COF            & CH$_4$ uptake (high P)                      & 69840& MAE     & 2.17e$+$01$^b$                  & 2.95e$+$00 & {\bf 2.35e$+$00} \\
     &                    & CH$_4$ uptake (low P)                       & 69840& MAE     & 6.63e$+$00$^b$                  & 1.20e$+$00 & {\bf 9.40e$-$01} \\
\hline 
PPNs & hPPN               & CH$_4$ uptake (1 bar)                       & 17846& MAE     & 1.23e$+$00$^b$                      & 7.21e$-$01 & {\bf 4.55e$-$01} \\
     &                    & CH$_4$ uptake (65 bar)                      & 17846& MAE     & 7.70e$+$00$^b$                  & 3.52e$+$00 & {\bf 3.09e$+$00} \\
\hline
ZEOs & ZEO--H             & Henry constant (low P)                      & 216  & MAE     & 2.87e$-$01$^b$                  & 2.26e$-$01 & {\bf 2.18e$-$01} \\
     &                    & Max loading (403 bar)                       & 216  & MAE     & 4.75e$+$00$^b$                  & 2.13e$+$00 & {\bf 2.19e$+$00} \\
\hline\hline
\end{tabular}%
}
\vspace{0.5em}
\begin{minipage}{\textwidth}
\footnotesize
$^a$ Bolded values indicate the best performance for each corresponding dataset or task.
$^b$ For datasets without reported source results, the CGCNN model is used as the baseline, with settings provided in Supplementary Information Section S3.2. 
$^c$ The text-mining labels were applied in the source paper\cite{nandy2021using}.
$^d$ The predefined train/validation/test splits from the source paper were used \cite{nandy2021using}.
\end{minipage}
\end{table}

We conducted a comprehensive evaluation of ITT on 18 downstream tasks covering adsorption, transport, stability, and electronic properties across MOFs, COFs, PPNs, and zeolites. When repeatable prior ML results were unavailable, we used CGCNN as a baseline under identical settings (details in Supplementary Information S3.2). For regression tasks we report MAE, RMSE, and $R^2$, and for classification we report Accuracy, Precision, and Recall. Aggregate outcomes are summarized in Table~\ref{tab:downstream_scratch_vs_itt_mae} (full metrics in Supplementary Information Table S6).

Against prior work and baselines, the proposed ITT achieves state-of-the-art results on 17 of 18 tasks under the same evaluation protocol. The only nuanced cases involve labels mined from the literature (SRS and TST), which can embed non-structural signals that structure-only models do not consistently capture. Even so, ITT matches or exceeds prior reports in most settings: for example, ITT improves thermal-stability MAE on the source fixed split, while on SRS it is slightly below the reported fixed-split accuracy but surpasses the scratch baseline on the randomized split (Accuracy = 0.77).

Relative to the ITT scratch variant (no pretraining), self-supervised pretraining delivers consistent gains across medium/large datasets—particularly for adsorption and transport (all eight O$_2$/N$_2$ tasks, MC-COF and hPPN methane uptake, CO$_2$ Henry). Two small gaps remain: ZEO–H max loading (403 bar), which is a small (~200) sample set where estimates are naturally high variance, shows virtually tied performance (MAE$_\text{ITT}$=2.19 vs. MAE$_\text{scratch}$=2.13), and QMOF-BD bandgap is marginally better with scratch in our runs. This behavior is consistent with the physics: bandgap is dominated by localized electronic environments, whereas our pretraining emphasizes global structural regularities; nevertheless, ITT’s multilevel design (atomic, elemental, interaction-topological, structural) retains strong local cues and remains highly competitive.

Representative scatter plots for N$_2$ self-diffusion at 1 bar and bandgap (Figure~\ref{fig:results_analysis}{\bf d,e}) show tight alignment with ground truth (MD/DFT), indicating good calibration across both physical (transport) and electronic properties. Token-level saliency (Figure~\ref{fig:results_analysis}{\bf f}) provides a mechanistic view: for a representative MOF (ID: \texttt{ABAXUZ\_FSR}), atomic- ($\sim$3 2\%) and elemental- ($\sim$ 33\%) level embeddings dominate bandgap prediction (blue), with interaction-topological ($\sim$ 19\%) features next—consistent with local electronic structure; in contrast, transport relies more on structural/elemental/interaction-topological signals that encode pathways at larger scales (red). This pattern is further confirmed by the distribution of token type importance across dataset samples (Figure~\ref{fig:results_analysis}{\bf g}): band gap prediction exhibits high reliance on localized features (followed by elemental-level, interaction topological embeddings), while transport-related tasks favor global or intermediate-scale representations. It is also noticed at Figure~\ref{fig:results_analysis}{\bf d} that a small number of obvious outliers in the bandgap scatter dominate the residual tail, it motivates us report all test samples under the prescribed splits and all downstream datasets are provided and also the testing set ids. Together, ITT’s self-supervised pretraining improves accuracy broadly, its multilevel embeddings adapt across disparate property types, and its saliency analysis yields interpretable attributions that align with known physics—local cues for electronic properties and broader structural cues for transport/adsorption.

\section{Discussion}

  \begin{figure}[!ht]
    \centering
    \includegraphics[width=16cm]{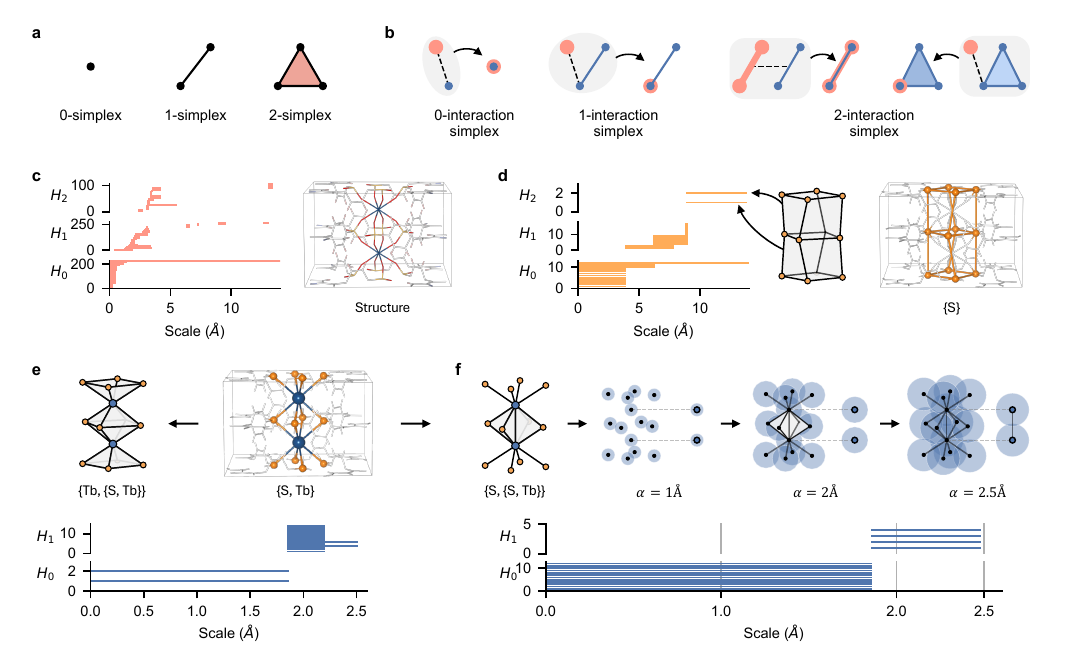}
    \caption{Interaction topology-based analysis.
    {\bf a} Building blocks of a simplicial complex, including 0-simplex (vertex), 1-simplex (edge), and 2-simplex (filled triangle).
    {\bf b} Building blocks of the interaction complex (n = 2), which generalizes simplicial complexes to describe interactions between two systems. 0-interaction simplex: intersection of two 0-simplices (vertices); 1-interaction simplex: intersection of a 0-simplex and a 1-simplex; 2-interaction simplices: formed either by the interaction of two 1-simplices or by the intersection of a 0-simplex with a 2-simplex. Pink and blue colors denote simplices from different systems.
    {\bf c} Persistent homology (PIH, $n$ = 1) in dimensions 0 to 2 for a MOF structure (ID: ABAXUZ\_FSR), capturing the global topological features of the full structure.
    {\bf d} Cluster-specific persistent homology (PIH, $n$ = 1) barcodes for the S atoms in the same MOF structure. In dimension 2 ($H_2$), two prominent bars represent two cube-like cavities defined by S atoms arranged in octameric rings.
    {\bf e} PIH ($n$ = 2) refers to the interaction between the cluster containing S atoms and the cluster containing Tb atoms, where Tb is the interaction center. $H_0$ captures the number of Tb atoms and their local connectivity with nearby S atoms, whereas $H_1$ captures loop-like triangular structures around Tb formed within the interaction complex.
    {\bf f} PIH ($n$ = 2) results for the S-Tb interaction, centered at S. Three alpha interaction complexes at increasing filtration radii ($\alpha$ = 1 \AA, 2 \AA, 2.5 \AA) illustrate the multiscale construction of interaction features. The persistent $H_1$ bars reflect triangle-like loops formed between S and neighboring Tb atoms, while $H_0$ bars capture the number of distinct S centers contributing to the interaction complex.
    }
    \label{fig:discussion_analysis}
  \end{figure}

To better capture the structural and chemical complexity of porous materials, the proposed ITT framework incorporates the interaction topology, e.g., PIH. As illustrated in Figure~\ref{fig:discussion_analysis}, PIH generalizes classical persistent homology (PH) by introducing interaction-aware topological analysis, enabling the extraction of both global and local structural features across multiple atomic subsystems. Traditional graph-based representations have been widely used in materials modeling, yet they are fundamentally limited to pairwise atomic interactions and lack the expressiveness to capture higher-order relationships. In contrast, simplicial complexes extend beyond pairwise interactions by using higher-dimensional building blocks, i.e., 0-, 1-, and 2-simplices, to represent atoms, bonds, and triangular relationships, respectively (Figure~\ref{fig:results_analysis}{\bf a}). Classical PH, when applied to such complexes, captures the overall topology of entire structures (e.g., connectivity, loops, voids), as demonstrated in Figure~\ref{fig:discussion_analysis}{\bf c}. However, its focus on the global structure often overlooks critical element-specific and interaction-level details necessary for understanding chemically heterogeneous systems. To incorporate chemical specificity, PIH ($n$ = 1) can be applied to elemental subsets of atoms. For example, Figure~\ref{fig:discussion_analysis}{\bf d} shows the persistent barcodes derived from sulfur (S) atoms in a MOF structure, where two prominent $H_2$ bars represent cube-like cavities bounded by octameric S rings. While this provides localized chemical insight, treating each element subset independently may still fail to capture inter-element interactions, which are especially important in multi-component systems like MOFs and COFs. Similarly, emerging classes of materials such as POM-MOFs, which feature cavity-centered POM units embedded in larger frameworks \cite{du2014recent}, represent ideal candidates for PIH analysis due to the interplay between their internal cluster topology and the surrounding MOF network.

To address this, we introduce the concept of interaction complexes (ICs), shown in Figure~\ref{fig:results_analysis}{\bf b}, which serve as the foundation for PIH with $n$ = 2. This approach considers interactions between pairs of atomic groups by constructing joint simplicial structures based on their geometric overlap. For example, Figure~\ref{fig:discussion_analysis}{\bf e, f} demonstrates how interactions between S and Tb atoms can be modeled from two perspectives—each using one element as the interaction center. When Tb is designated as the center (Figure~\ref{fig:discussion_analysis}{\bf e}), $H_0$ captures the number and connectivity of Tb atoms with nearby S atoms, while $H_1$ reflects loop-like motifs formed in the vicinity of Tb. Conversely, centering on S atoms (Figure~\ref{fig:discussion_analysis}{\bf f}) reveals different patterns: $H_0$ tracks distinct S-centered interaction sites, and $H_1$ identifies triangular loops involving neighboring Tb atoms. The construction of interaction complexes at multiple filtration scales (e.g., $\alpha$ = 1, 2, 2.5 \AA) highlights the multiscale nature of PIH, which can systematically capture geometric and topological transitions across spatial resolutions. 
{ In the context of POM-MOFs, this multiscale capability could be especially valuable for resolving cavity-specific interactions between POM clusters and the surrounding MOF scaffold, enabling a detailed topological characterization of these hybrid materials.}

PIH’s strength lies in its ability to integrate interaction-centered perspectives with topological analysis, preserving both fine-grained local interactions and larger-scale structural organization. Unlike traditional PH, which is structure-centric, PIH enables analysis of inter-system relationships, a critical capability for modeling functional behavior in chemically diverse materials. This dual-level characterization—distinguishing not only the topology within clusters but also the interactions between them—makes PIH a powerful foundation for the multi-level embeddings used in ITT. Ultimately, PIH enriches the ITT framework by providing topological features that encode both compositional and relational information, allowing the model to reason over the structural hierarchy inherent in real-world material systems.

{ Looking forward, the ITT framework holds particular promise for advancing the design and discovery of POM-based materials, such as POM-MOFs and polymer-modified POMs. These systems are especially interesting because POMs act as cavity-centered clusters whose functional properties, such as redox activity and charge states, are strongly modulated by their chemical surroundings (e.g., ligands, ordered MOF scaffolds, or polymer matrices) \cite{yang2023old}. The multiscale capability of ITT is well suited for this challenge: atomic-level embeddings can capture charge-sensitive features, element- and pairwise-element embeddings from interaction topology can describe POM–ligand or POM–scaffold interactions, and structure-level embeddings can resolve the influence of long-range crystalline environments. Together, these representations could guide the rational design of POM-based materials with tailored electronic structures and memristive states, offering a systematic pathway for extending ITT to functional materials beyond conventional porous frameworks.}

\section{Method}

\subsection{Data Collection}

In this work, we assembled a multi-domain corpus of porous materials, including MOFs, COFs, PPNs, and zeolites (ZEOs), to support structure-only self-supervised pretraining and supervised fine-tuning of the ITT model. For pretraining, we used five complementary structure collections strictly as unlabeled crystals: ARC-MOF \cite{burner2023arc}, Materials Cloud COFs (MC-COF) \cite{mercado2018silico}, CoRE-COF (v7) \cite{tong2017exploring}, amorphous PPNs \cite{park2024large}, and IZA–SC zeolites \cite{baerlocher2007atlas}. ARC-MOF aggregates experimentally characterized and hypothetical MOFs; MC-COF comprises in-silico COFs built from a large linker library and relaxed for simulation; CoRE-COF (v7) provides curated experimental COF structures; the PPN set contains amorphous polymer networks generated via self-assembly simulations; and IZA–SC hosts canonical zeolite framework structures. In total, the pretraining pool includes 602,396 structures and can be readily expanded with large, label-free structure repositories. For fine-tuning, we assembled labeled datasets across MOFs, COFs, PPNs, and zeolites, using an 8:1:1 train/validation/test random split for every task. The MOF suite includes the eight-task $\mathrm{N_2}/\mathrm{O_2}$ selectivity collection derived from the simulated CoRE MOF 2019 database \cite{chung2019advances} (see Supplemental Table S7); to improve robustness, we removed extreme outliers using the upper-limit thresholds introduced by \citet{orhan2021prediction}, yielding a more uniform target-variable distribution. Additional MOF labels include Solvent-Removal Stability (binary activation stability) and Thermal Stability (decomposition temperatures) mined from the literature via text and image analysis \cite{nandy2021using}, QMOF band gaps computed by periodic DFT workflows \cite{rosen2021machine}, and a CO$_2$ Henry-coefficient dataset reported as $\log k_H$ \cite{park2023enhancing}. For COFs, we used the MC-COF dataset with methane-uptake labels generated by grand canonical Monte Carlo simulations \cite{mercado2018silico}; for PPNs, the hPPN table with simulated methane-adsorption labels \cite{martin2014silico}; and for zeolites, the Zeolite\_Hydrogen (ZEO–H) subset with labels (low-pressure slope and high-pressure loading) produced by high-throughput Monte Carlo hydrogen-adsorption simulations \cite{sun2021fingerprinting}. To ensure fair and comparable evaluation, we did not introduce any additional designed structures that could bias the model during either pretraining or fine-tuning; all datasets were collected from publicly available, fully traceable sources. Details on dataset sizes and brief descriptions are summarized in Table~\ref{tab:datasets}.

\begin{table}[htbp]
\centering
\caption{Summary of porous materials datasets.}
\label{tab:datasets}
\resizebox{0.98\textwidth}{!}{
\begin{tabular}{p{2.2cm} p{3cm} p{2cm} l}
\hline
\textbf{Purpose} & \textbf{Dataset} & \textbf{Size} & \textbf{Properties Used} \\
\hline
Pretraining & ARC-MOF \cite{burner2023arc}           & 520,835 & MOF structures only \\
            & MC-COF \cite{mercado2018silico}           & 69,840  & COF structures only (Materials Cloud) \\
            & CoRE-COF \cite{tong2017exploring}      & 1,242   & COF structures only (CoRE-COF version 7) \\
            & PPN \cite{park2024large}               & 10,237  & Amorphous PPN structures only \\
            & IZA-SC \cite{baerlocher2007atlas}      & 242     & Zeolite structures only \\
\hline
Downstream (MOFs) & O\textsubscript{2}/N\textsubscript{2}-Select \cite{chung2019advances,orhan2021prediction} & 4,744--5,241 & Henry constants, uptakes, and self-diffusivities for O\textsubscript{2}/N\textsubscript{2} (8 tasks; see Table S7) \\
                 & SRS \cite{nandy2021using}       & 2,179 & Solvent-removal stability (binary classification) \\
                 & TST \cite{nandy2021using}        & 3,132 & Thermal stability temperature (K) \\
                 & QMOF-BD \cite{rosen2021machine}    & 20,375 & Bandgap (eV) \\
                 & CO\textsubscript{2}--Henry \cite{park2023enhancing} & 9,525 & CO\textsubscript{2} Henry constant (log $k_H$; $k_H$ in mol$\cdot$kg$^{-1}\cdot$Pa$^{-1}$) \\
\hline
Downstream (COFs) & MC-COF \cite{mercado2018silico}       & 69,840  & CH\textsubscript{4} uptake at 65 bar and 5.8 bar (cm$^{3}$(STP)$\cdot$cm$^{-3}$) \\
\hline
Downstream (PPNs) & hPPN \cite{martin2014silico}                   & 17,846  & CH\textsubscript{4} uptake at 1 bar and 65 bar (cm$^{3}$(STP)$\cdot$cm$^{-3}$) \\
\hline
Downstream (ZEOs) & ZEO--H \cite{sun2021fingerprinting}   & 215     & Henry constant at low $P$ ((g/L)/bar), max loading at 403 bar (g/L) \\
\hline
\end{tabular}
}
\end{table}

\subsection{Interaction Topology Theory}
\subsubsection{Interaction complex}
To incorporate element-specific information when analyzing porous materials, we partition atoms into elemental clusters and study how these clusters jointly form topological features. At a fixed scale, let $K$ be the simplicial complex on all atoms, and let $\{K_i\}_{i=1}^n$ be a family of subcomplexes of $K$ induced by chosen element clusters (e.g., metals, oxygens, carbons). An interaction simplex is a tuple $(\sigma_1,\ldots,\sigma_n)$ with $\sigma_i\in K_i$ whose intersection is nonempty,
$$
\sigma_1\cap \cdots \cap \sigma_n\neq\varnothing,
$$
representing a joint feature contributed by all selected clusters. The demonstrations for 0-, 1-, and 2-interaction simplex are shown in Figure~\ref{fig:discussion_analysis}{\bf b}. The resulting $n$-interaction complex is
$$
\mathcal K=\big(K,\{K_i\}_{i=1}^n\big).
$$
Specifically, when ${n=1}$, $\mathcal K=(K,\{K\})$ reduces to the ordinary simplicial complex $K$. We use this in two ways: (i) structural-level embedding on all atoms (treating atoms as identical), and (ii) elemental-level embedding by applying the same construction to a single cluster $K_i$ (capturing element-specific structure). When ${n=2}$, $\mathcal K=(K,\{K_1,K_2\})$ captures pairwise interactions between two element clusters. For convenience, we write $\{K_i\}$ in place of $\big(K,\{K_i\}_{i=1}^n\big)$ whenever no confusion arises.

\subsubsection{Interaction homology and its persistence}
Analogous to ordinary homology, we form interaction chains by taking formal sums of interaction simplices and use a boundary operation that acts on one component at a time (the usual simplicial boundary applied componentwise while keeping the other components fixed). This identifies cycles (joint features with no boundary) and boundaries (those arising from higher-dimensional joint features). The resulting interaction homology group measures $p$-dimensional joint features—shared connected components ($p=0$), co-supported loops/tunnels ($p=1$), and co-supported cavities ($p=2$) that exist because specific element clusters arrange together. The rank of the interaction homology group is the interaction Betti number
$$
\beta_p \;=\; \dim_{\mathbb K}\, H_p(\{K_i\};\mathbb K).
$$
Here, $\mathbb{K}$ is the base field, and $H_p(\{K_i\};\mathbb K)$ denotes the $p$-dimensional interaction homology of $\{K_i\}$.
To capture how these joint features change with spatial scale, we use persistent interaction homology. As the scale parameter $a$ increases, both the whole complex $K(a)$ and each subcomplex $K_i(a)$ grow, producing a nested family of interaction complexes $\{K_i(a)\}$. For real numbers $a\le b$, the inclusion $\{K_i(a)\}\to \{K_i(b)\}$ induced by $a\to b$ gives a map on the interaction homology. The $(a,b)$-persistent interaction homology group and its persistent Betti number are
$$
H_p^{a,b}(\mathcal{F}_i;\mathbb K)
=\operatorname{im}\!\Big(
H_p\big(\mathcal F_i(a)\big)\xrightarrow{\;\;}
H_p\big(\mathcal F_i(b)\big)
\Big),
\qquad
\beta_p^{a,b} \;=\; \dim_{\mathbb K}\, H_p^{a,b}(\mathcal F;\mathbb K),
$$
where $\mathcal F(a)=(K(a),\{K_i(a)\}_{i=1}^n)$. Long-lived (persistent) joint features correspond to stable, chemically meaningful patterns, e.g., channels framed by metals and linkers—whereas short-lived ones are often noise or highly localized artifacts. In this work, the alpha complex is used to represent a single system, while the interactions between alpha complexes are denoted by the alpha interaction complex (alpha-IC), as shown in Figure~\ref{fig:discussion_analysis}{\bf f}.

If there is a single cluster ($n=1$ and $K_1=K$), the interaction homology reduces to the ordinary homology, and the persistent interaction homology reduces exactly to the classical persistent homology. In this work, the structural embedding corresponds to $n=1$ on all atoms (Figure~\ref{fig:discussion_analysis}{\bf c}); the elemental embedding corresponds to $n=1$ on each cluster $K_i$ (Figure~\ref{fig:discussion_analysis}{\bf d}); and the interaction embedding uses $n=2$ to quantify features arising specifically from the interaction of two element clusters, i.e., $H_p^{a,b}((K,\{K_1, K_2 \});\mathbb K)$, providing element-aware, multiscale descriptors beyond what either cluster shows alone (Figure~\ref{fig:discussion_analysis}{\bf e, f}). The detail and rigorous formulation of interaction topology are provided at the supplementary information Note 4.

\subsection{Predictive Modeling}

\subsubsection{Data Preprocessing}
In this work, each material’s supercell is uniformly scaled to approximately $64 \, \text{\AA} \times 64 \, \text{\AA} \times 64 \, \text{\AA}$ to ensure consistent topological analysis across structures. For the structural- and elemental-level embeddings, we employ the HIP method with $n=1$ (persistent homology) to compute topological invariants in the $H_0$, $H_1$, and $H_2$ homology dimensions. Those topological embeddings are generated using a grid-based approach: a grid from $0$ to $25 \, \text{\AA}$ with a step size of $0.1 \, \text{\AA}$ is applied, and the Betti numbers—representing the number of topological features at each scale—are recorded. This produces a feature vector of length 750 ($250$ grid steps $\times$ 3 homology dimensions). The structural-level embedding consists of a single token (shape $1 \times 750$), while the elemental-level embedding contains seven tokens ($7 \times 750$). Interaction-level topological embeddings are computed using the same grid for the $H_0$ and $H_1$ interaction homology dimensions, yielding a 500-dimensional feature vector for each elemental interaction pair. With seven elemental clusters, there are 42 distinct interaction pairs, resulting in $42 \times 500$ features in total. For atomic-level embeddings, duplicate atoms are removed by comparing local neighbor environments. Each remaining atom is initialized with features following the CGCNN scheme, using an $8 \, \text{\AA}$ cutoff to identify neighbors.

\subsubsection{Computational settings}

The IIT model consists of eight hidden layers (ITT decoders) with a hidden size of 256. The multi-head attention modules use eight heads, and the feed-forward layers have an intermediate size of 1024. For the atomic embeddings, four convolutional layers are applied, each producing atomic embeddings of size 128. During pretraining, the model is trained for 200 epochs with a batch size of 256 and the random mask rate 50\% for the first token. For finetuning, a batch size of 32 is used for all downstream tasks, and training is run for 200 epochs. In both pretraining and finetuning, the AdamW optimizer is employed with a weight decay of 0.01 and a base learning rate of 0.0001. A warm-up learning rate schedule is applied, where the first 10\% of the total epochs are designated as warm-up. Mean squared error is used as the loss function for regression tasks, and cross-entropy is used for classification tasks. To stabilize model performance, all labels are standardized before training to mitigate issues arising from extremely large value ranges. Additionally, for each downstream task, the predictions from the top five models (ranked by validation performance during training) are averaged for each sample, resulting in more stable and robust outputs. Further details on the hyperparameters and model formulations are provided in Table S2-S4 and Supplementary Note 3.

\section{Data availability}
All porous-materials datasets analyzed in this study are publicly available from the sources cited in the manuscript. The curated bundle used in our studies, including strucature files and target labels, is trackable at \url{https://github.com/WeilabMSU/ITTransformer} to facilitate full reproducibility.

\section{Code availability}
The source code and model implementations used in this work are openly available at \url{https://github.com/WeilabMSU/ITTransformer}. 

\section{Acknowledgments}
This work was supported in part by NIH grant R35GM148196, National Science Foundation grant DMS2052983,  Michigan State University Research Foundation, and  Bristol-Myers Squibb 65109. D. C. was supported in part by the AMS-Simons Travel Grant.
C.-L.C. gratefully acknowledges financial support from the Defense Threat Reduction Agency (Project CB11141), and the Department of Energy (DOE), Office of Science, Office of Basic Energy Sciences (BES) under an award FWP 80124 at Pacific Northwest National Laboratory (PNNL). PNNL is a multiprogram national laboratory operated for the Department of Energy by Battelle under Contract DE-AC05-76RL01830.
C.-L. C. and G.-W. W would like to also acknowledge financial support from DOE, Office of Science, BES under an award FWP 84274 for the development of machine learning models.

\newpage
\bibliographystyle{unsrtnat}  

\end{document}